\documentclass[letterpaper]{article} 
\usepackage{aaai24}  
\usepackage{times}  
\usepackage{helvet}  
\usepackage{courier}  
\usepackage[hyphens]{url}  
\usepackage{graphicx} 
\usepackage{natbib}  
\usepackage{caption} 
\usepackage{algorithm}
\usepackage{algorithmic}

\usepackage{newfloat}
\usepackage{listings}
\DeclareCaptionStyle{ruled}{labelfont=normalfont,labelsep=colon,strut=off} 
\floatstyle{ruled}
\newfloat{listing}{tb}{lst}{}
\floatname{listing}{Listing}
\usepackage{booktabs}
\usepackage{tabularx}
\usepackage{multirow}
\usepackage{amsthm}
\usepackage{amsmath}
\usepackage{amssymb}
\usepackage{utfsym}

\title{MmAP : Multi-modal Alignment Prompt for Cross-domain Multi-task Learning}
\author{
    Yi Xin\textsuperscript{\rm 1,2}\equalcontrib,
    Junlong Du\textsuperscript{\rm2}\equalcontrib,
    Qiang Wang\textsuperscript{\rm2},
    Ke Yan\textsuperscript{\rm2}\thanks{Corresponding author.},
    Shouhong Ding\textsuperscript{\rm2}
}
\affiliations{
    \textsuperscript{\rm 1}State Key Laboratory for Novel Software Technology, Nanjing University, Nanjing, China\\
    \textsuperscript{\rm 2}Youtu Lab, Tencent\\
    xinyi@smail.nju.edu.cn, jeffdu@tencent.com, albertqwang@tencent.com, \\kerwinyan@tencent.com, ericshding@tencent.com 
}

\usepackage{bibentry}

\begin{document}

\maketitle

\begin{abstract}
Multi-Task Learning (MTL) is designed to train multiple correlated tasks simultaneously, thereby enhancing the performance of individual tasks. Typically, a multi-task network structure consists of a shared backbone and task-specific decoders. However, the complexity of the decoders increases with the number of tasks. To tackle this challenge, we integrate the decoder-free vision-language model CLIP, which exhibits robust zero-shot generalization capability. Recently, parameter-efficient transfer learning methods have been extensively explored with CLIP for adapting to downstream tasks, where prompt tuning showcases strong potential. Nevertheless, these methods solely fine-tune a single modality (text or visual), disrupting the modality structure of CLIP. In this paper, we first propose \textbf{M}ulti-\textbf{m}odal \textbf{A}lignment \textbf{P}rompt (\textbf{MmAP}) for CLIP, which aligns text and visual modalities during fine-tuning process. Building upon MmAP, we develop an innovative multi-task prompt learning framework. On the one hand, to maximize the complementarity of tasks with high similarity, we utilize a gradient-driven task grouping method that partitions tasks into several disjoint groups and assign a group-shared MmAP to each group. On the other hand, to preserve the unique characteristics of each task, we assign an task-specific MmAP to each task. Comprehensive experiments on two large multi-task learning datasets demonstrate that our method achieves significant performance improvements compared to full fine-tuning while only utilizing approximately $\sim0.09\%$ of trainable parameters.
\end{abstract}

\section{Introduction}
\label{sec:intro}
\begin{figure}[t]
   \begin{picture}(0,227)
     \put(-5,-5){\includegraphics[width=1\linewidth]{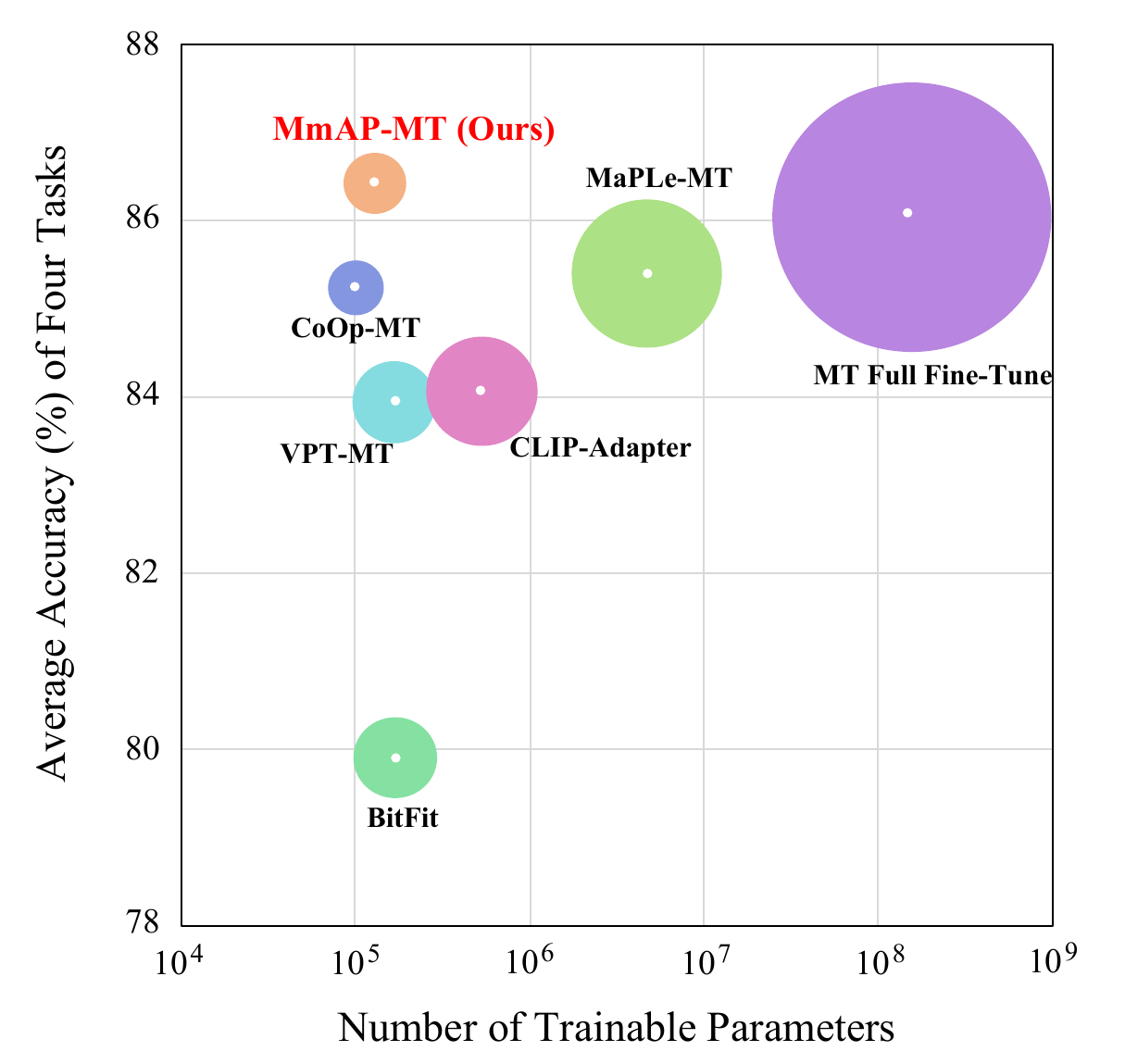}}
   \end{picture}
    \caption{The trade-off between average accuracy over four tasks on Office-Home~\cite{venkateswara2017deep} dataset and the number of trainable parameters. The radius of each circle represents the relative amount of trainable parameters.} 
    \label{fig: pipeline}
\end{figure}
\begin{figure*}
	\centering
 	\includegraphics[width=1\textwidth]{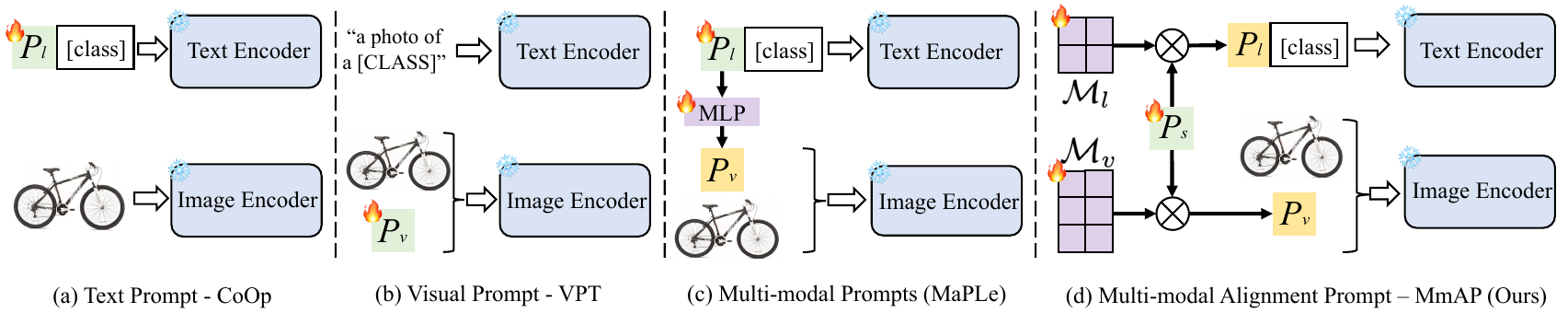}
	\caption{Illustrations of (a) text prompt tuning~\cite{zhou2022learning}, (b) visual prompt tuning~\cite{jia2022vpt}, (c) multi-modal prompts learning~\cite{khattak2023maple} and (d) our multi-modal alignment prompt tuning. \includegraphics[width=0.32cm]{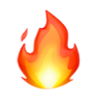} represents trainable parameters, \includegraphics[width=0.32cm]{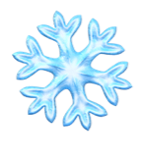} represents frozen parameters, $\otimes$ represents Kronecker Product and [class] represents category name.}
	\label{fig: prompt}
\end{figure*}
Multi-Task Learning (MTL) has surfaced as a potent approach in deep learning that allows for joint training of multiple correlated tasks within a unified network architecture, resulting in enhanced model performance in comparison to Single-Task Learning (STL). The core of MTL lies in learning both the task-shared and the task-specific representations. By capitalizing on shared representations and knowledge across tasks, MTL enhances generalization and mitigates overfitting. Utilizing specific representations allows MTL to preserve the distinct characteristics of each task. 
Moreover, training a unified model for multiple tasks is generally more parameter-efficient than training several single-task models. Consequently, MTL has garnered considerable interest in various fields, including Computer Vision~\cite{shen2021variational,ye2022taskprompter}, Natural Language Processing~\cite{he2022hyperprompt}, etc.


In this work, we mainly focus on vision multi-task learning. Prior research has predominantly concentrated on the design of multi-task model training framework, encompassing encoder-based methods~\cite{gao2019nddr} and decoder-based methods~\cite{xu2023demt}. However, with the growing prowess of vision pre-trained models (e.g., ViT~\cite{dosovitskiy2020image}, SwinTransformer~\cite{liu2021swin}), directly fine-tuning these models for downstream multi-task leads to substantial performance enhancements and has become the mainstream approach for multi-task learning~\cite{liu2022polyhistor}. In this fine-tuning paradigm, it remains necessary to establish a distinct decoder for each task, with trainable parameters that increase linearly.

To address the above issue, we incorporate the pre-trained vision-language model CLIP~\cite{radford2021learning} and consider it tailor-made for vision multi-task learning. On one hand, CLIP is trained to align language and vision modalities using web-scale data (e.g., 400 million text-image pairs), endowing it with a robust capability for zero-shot transfer to vision downstream tasks. On the other hand, the architecture of CLIP offers a distinct advantage. It comprises a text encoder and an image encoder, eliminating the need to establish additional decoder structures for each task. Therefore, we opt for adapting CLIP to address vision multi-task.

Following the conventional pretrain-finetune paradigm, the entire CLIP parameters ($\sim$150M) would require updating, which presents challenges concerning computational and storage expenses. Recently, numerous studies~\cite{zaken2021bitfit, jia2022vpt, gao2021clip, zhou2022learning} have introduced parameter-efficient transfer learning techniques to achieve an optimal balance between trainable parameters and performance on downstream tasks. Nonetheless, these existing methods primarily concentrate on pre-trained vision models or language models, with their applicability to more complex vision-language models remaining uncertain. Moreover, these approaches tend to emphasize single-task adaptation, while multi-task adaptation continues to pose a challenge.

To start with, we initially conduct a thorough examination of the performance of existing successful parameter-efficient transfer learning methods when applied to CLIP for vision multi-task learning, as shown in Figure~\ref{fig: pipeline}. Through our extensive studies, we discover that prompt tuning methods VPT-MT \cite{jia2022vpt}, CoOp-MT \cite{zhou2022learning} and MaPLe-MT~\cite{khattak2023maple} are more suitable than BitFit \cite{zaken2021bitfit} and Adapter \cite{gao2021clip}. This may be attributed to the fact that BitFit and Adapter update model parameters and disrupt the original structural integrity of CLIP. In contrast, prompt tuning methods only modifies input embedding (text or image), as shown in Figure~\ref{fig: prompt}. Moreover, we observe that MaPLe-MT outperforms VPT-MT and CoOp-MT, emphasizing the advantages of tuning both modalities simultaneously.

Subsequently, based on our observations, we propose a novel Multi-modal Alignment Prompt (MmAP) for CLIP along with a framework tailored for multi-task image recognition scenarios. Our MmAP generates text prompts and visual prompts through a source prompt to achieve the tuning alignment effect for both modalities. Additionally, we design a multi-task prompt tuning framework based on MmAP. Previous MTL works~\cite{fifty2021efficiently, standley2020tasks} have confirmed that training similar tasks together yields a complementary effect, while training dissimilar tasks together results in a negative effect. Therefore, we first employ gradient similarity to group tasks and then assign a group-shared MmAP for joint training. Furthermore, to maintain the independent characteristics of each task, we establish task-specific MmAP for each task individually. We evaluate our method on two large cross-domain multi-task datasets, including Office-Home and MiniDomainNet. Figure~\ref{fig: pipeline} displays the results on Office-Home, illustrating that our proposed method has achieved a favorable trade-off between trainable parameters and performance.

Our main contributions are as follows:
\begin{itemize}
    \item We propose \textbf{M}ulti-\textbf{m}odal \textbf{A}lignment \textbf{P}rompt (\textbf{MmAP}) for CLIP to favourably align its vision-language representations while parameter-efficient tuning.
    \item Building upon MmAP, we design a multi-task prompt learning framework for cross-domain image recognition tasks, incorporating both group-shared MmAP and task-specific MmAP.
    \item We devise a unified library grounded in CLIP to benchmark various parameter-efficient tuning methods for multi-task image recognition. To the best of our knowledge, we are the first to undertake this work.
    \item Experimental results on two commonly used visual multi-task datasets show that our method achieves competitive performance compared to multi-task full fine-tuning leveraging merely $\sim0.09\%$ of the CLIP parameters, as shown in Figure~\ref{fig: pipeline}.
\end{itemize}

\section{Related Work}
\label{sec:relate}
\paragraph{Multi-Task Learning.} 
Multi-Task Learning (MTL) aims to simultaneously learn multiple tasks by sharing knowledge and computation. There are two classic multi-task in the field of computer vision. The first is dense scene understanding multi-task, which implements semantic segmentation, surface normal estimation, saliency detection, etc. for each input sample. Current research on multi-task dense scene understanding primarily focuses on decoder structure design~\cite{ zhang2021transfer,xu2023demt, liang2023visual}. The other is cross-domain classification multi-task, and the input data consists of multiple datasets with domain shifts. As multiple domains are involved, current research emphasizes learning shared and private information between domains~\cite{shen2021variational, long2017learning}.

\paragraph{Vision-Language Model.} Foundational vision-language models (e.g., CLIP~\cite{radford2021learning} and  ALIGN~\cite{jia2021scaling}) have exhibited remarkable capabilities in various vision tasks.  In contrast to models learned with only image supervision, these V-L models encode rich multimodal representations. Although these pre-trained V-L models learn rich representations, efficiently adapting them to downstream vision tasks remains a challenging problem. Numerous works have demonstrated improved performance on downstream vision tasks by employing tailored methods to adapt V-L models for detection~\cite{li2022grounded, zhong2022regionclip}, segmentation~\cite{rao2022denseclip, xu2022groupvit}, and recognition~\cite{wortsman2022robust}. Furthermore, HiPro~\cite{liu2023hierarchical} constructs a hierarchical structure to adapt a pre-trained V-L model to various downstream tasks.

\paragraph{Parameter-Efficient Transfer Learning.} 
Parameter Efficient Transfer Learning (PETL) aims to adapt a pre-trained model to new downstream tasks by training only a small number of parameters. Existing PETL methods can be categorized into three groups: parameter tuning, adapter tuning, and prompt tuning. Parameter tuning directly modifies the parameters of a pre-trained model, either by tuning the weights~\cite{hu2021lora} or biases~\cite{zaken2021bitfit}. Adapter tuning inserts trainable bottleneck architectures into a frozen pre-trained model, intending to facilitate learning for downstream tasks, such as AdaptFormer~\cite{chen2022adaptformer}, VL-Adapter\cite{sung2022vladapter}, and CLIP-Adapter~\cite{gao2021clip}. Prompt tuning unifies all downstream tasks into pre-trained tasks via designing a specific template to fully exploit the capabilities of foundation models~\cite{jia2022visual, khattak2023maple, wang2023seeing}. 


%

\begin{figure*}[t]
    \centering
   \includegraphics[width=1\textwidth]{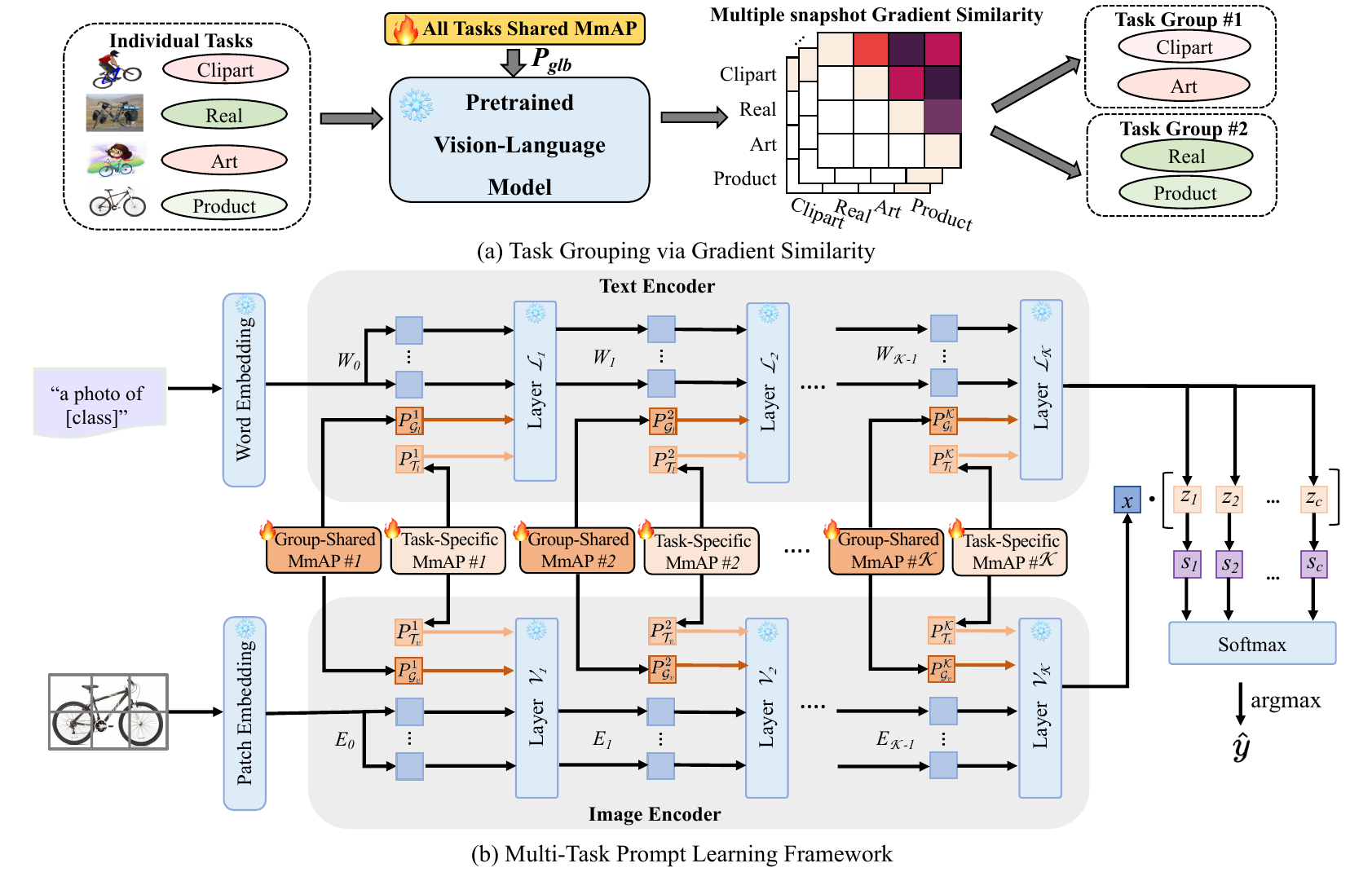}
    \caption{\textbf{Multi-Task Prompt Learning Framework}, including (a) grouping tasks by maximizing the complementarity of tasks with high similarity and (b) employing group-shared and task-specific MmAP for adapting CLIP to downstream tasks.}
    \label{fig: framework}
\end{figure*}
\section{Method}
\label{sec:method}
In this section, we first revisit vision-language models with a focus on CLIP. Subsequently, we introduce our proposed \textbf{M}ulti-\textbf{m}odal \textbf{A}lignment \textbf{P}rompt (\textbf{MmAP}).
Finally, we propose a unified prompt learning framework that incorporates both group-shared MmAP and task-specific MmAP.

\subsection{Contrastive Language-Image Pre-training}
\paragraph{Image Encoder.} In this work, we opt for ViT~\cite{dosovitskiy2020image} as the image encoder to be compatibile with the visual prompt~\cite{jia2022vpt}. 
Given an input image $I\in R^{H\times W \times 3}$, the image encoder, consisting of $\mathcal{K}$ transformer layers, splits the image into $M$ fixed-size patches and projects them into patch embeddings $E_{0}\in R^{M\times d_{v}}$. The patch embeddings $E_{k}$, accompanied by a learnable class token $c_{k}$, are fed into the $(k+1)$-th layer $\mathcal{V}_{k+1}$ of the image encoder, 
and sequentially processed through the following transformer layers:
\begin{equation}
    [c_{k+1}, E_{k+1}] = \mathcal{V}_{k+1}([c_{k}, E_{k}]) \quad k=0,1,...,\mathcal{K}-1.
\end{equation}
To acquire the ultimate image representation $x$, the class token $c_{\mathcal{K}}$ from the last transformer layer is projected into the V-L latent embedding space via \textit{ImageProj}:
\begin{equation}
    x = {\rm ImageProj}(c_{\mathcal{K}}).
\end{equation}
\paragraph{Text Encoder.} The text encoder adopts a transformer that contains $\mathcal{K}$ layers to tokenize the input words and project them into word embeddings $W_{0}\in R^{N\times d_{l}}$. The $W_{k}$ are directly fed into the $(k+1)$-th layer $\mathcal{L}_{k+1}$ of the text encoder:
\begin{equation}
    [W_{k+1}] = \mathcal{L}_{k+1}(W_{k}) \quad k=0,1,...,\mathcal{K}-1.
\end{equation}
The final text representation $z$ is obtained by projecting the text embeddings associated with the last token of the concluding transformer layer $\mathcal{L}_{\mathcal{K}}$ into V-L latent embedding space via \textit{TextProj}:
\begin{equation}
    z = {\rm TextProj}(W_{\mathcal{K}}).
\end{equation}

\paragraph{Zero Shot Prediction.} For zero-shot prediction, a carefully designed prompt is introduced into the language branch of CLIP, which serves to reconstruct the textual input by equipping with every class name associated with the downstream tasks (e.g., ``a photo of a [CLASS]''). The class with the highest cosine similarity score is then selected as the predicted label $\hat{y}$ for the given image, namely that:
\begin{equation}
    p(\hat{y}|x)=\frac{\exp \left(\operatorname{sim}\left(x, z_{\hat{y}}\right) / \tau\right)}{\sum_{i=1}^{C} \exp \left(\operatorname{sim}\left(x, z_{i}\right)/ \tau\right)},
\end{equation}
where $sim(·,·)$ represents the computation of cosine similarity. $\tau$ is the temperature coefficient learned by CLIP and $C$ is the total number of classes.

\subsection{Multi-modal Alignment Prompt}
Prior research has mainly focused on designing prompts for a single modality. For example, VPT investigated visual prompts, while CoOp introduced learnable text prompts. We believe that merely tuning one modality disrupts the text-image matching of CLIP, leading to sub-optimal adaptation for downstream tasks. 
The most concurrent method MaPLe proposed to use text prompt to generate visual prompt via a MLP with considerable parameters, which exhibits limitations regarding to visual modality and model efficiency.

To address these issues, we propose \textbf{M}ulti-\textbf{m}odal \textbf{A}lignment \textbf{P}rompt (\textbf{MmAP}) to generate text prompt $P_{l}\in R^{b\times d_{l}}$ and image prompt $P_{v}\in R^{b\times d_{v}}$ simultaneously, as depicted in Figure~\ref{fig: prompt}d. Here, we denote $b$ as the length of prompts, while $d_{l}$ and $d_{v}$ indicate the dimension of text and image tokens, respectively. We first initialize a source prompt $P_{s}\in R^{m\times n}$ and two individual scaling matrices $\mathcal{M}_{l}\in R^{\frac{b}{m} \times \frac{d_{l}}{n}}$ and $\mathcal{M}_{v}\in R^{\frac{b}{m} \times \frac{d_{v}}{n}}$ for two modalities. Then, we apply Kronecker Product to generate prompts for text and image encoders as follows: 
\begin{equation}    
\begin{aligned}
P_{l}=\mathcal{M}_{l} \otimes P_{s}=\left(\begin{array}{ccc}
\mathcal{M}_{l_{11}}P_{s} & \cdots & \mathcal{M}_{l_{1n}} P_{s} \\
\vdots & \ddots & \vdots \\
\mathcal{M}_{l_{m1}}P_{s} \mathbf{} & \cdots & \mathcal{M}_{l_{mn}}P_{s} \mathbf{}
\end{array}\right),
\end{aligned}
\end{equation}
\begin{equation}
\begin{aligned}
P_{v}=\mathcal{M}_{v} \otimes P_{s}=\left(\begin{array}{ccc}
\mathcal{M}_{v_{11}}P_{s} & \cdots & \mathcal{M}_{v_{1n}} P_{s} \\
\vdots & \ddots & \vdots \\
\mathcal{M}_{v_{m1}}P_{s} & \cdots & \mathcal{M}_{v_{mn}}P_{s}
\end{array}\right).
\end{aligned}
\end{equation}

Our proposed MmAP offers two significant advantages. Firstly, the use of the Kronecker Product ensures maximum preservation of the information of source prompt $P_s$. This facilitates alignment between the text and image prompts.
Secondly, the number of learnable parameters is significantly reduced from $\mathcal{K}(d_{l}+d_{v})$ to $mn+\frac{\mathcal{K}(d_{l}+d_{v})}{mn}$, where $\mathcal{K}$ represents the number of transformer layers. This reduction in parameters not only makes the model more efficient but also reduces the risk of overfitting. 

\subsection{Multi-Task Prompt Learning Framework}
In multi-task learning, the joint training of similar tasks can yield mutually beneficial outcomes. Typically, the degree of task similarity can be quantified by evaluating the gradient conflict between tasks. In light of this, we first involves grouping similar tasks together. A shared MmAP is assigned to each group which facilitates the mutual learning and enhancement of tasks within the group. However, to maintain the unique characteristics of each task, we also assign an individual MmAP to each task. This individual MmAP ensures that the distinct features and requirements of each task are adequately catered to. The overall multi-task prompt tuning framework diagram is depicted in Figure \ref{fig: framework}.

\paragraph{Task Grouping.} 
Existing MTL works~\cite{fifty2021efficiently} have demonstrated that gradient cosine similarity can quantify the similarity of two tasks, i.e., the extent to which two tasks can benefit from joint training. Therefore, we assess the similarity of two tasks by computing gradients on the shared parameters, while keeping the pretrained vision-language model frozen, as shown in Figure \ref{fig: framework}a.

Specifically, given a global shared MmAP $P_{glb}$ for all tasks, the similarity between the $i$-th task and the $j$-th task can be estimated as the following dot product:
\begin{equation}
    sim(\mathcal{T}_{i}, \mathcal{T}_{j})=\nabla_{P_{glb}} L_{\mathcal{T}_{i}}\left(P_{glb}\right) \cdot \nabla_{P_{glb}} L_{\mathcal{T}_{j}}\left(P_{glb}\right),
\end{equation}
where $L_{\mathcal{T}}$ denotes the loss on task $\mathcal{T}$. We posit that when $sim(\mathcal{T}_{i}, \mathcal{T}_{j})>0$, it indicates that the two tasks exhibit a mutual gain effect. Moreover, for robust estimation, we average multiple ``snapshots'' of similarity during the training of the global shared MmAP. At a high level, we concurrently train all tasks, evaluate pairwise task similarity throughout the training process, and identify task groups that maximize the total inter-task similarity.

\paragraph{Multi-Task Prompt Learning.} 
We develop a unified multi-task prompt learning framework upon our proposed MmAP, as depicted in Figure~\ref{fig: framework}b. Given $N$ downstream tasks $\{\mathcal{T}_i\}_{i=1}^{N}$, we first partition them into several disjoint groups according to gradient similarities. For brevity, we denote $\mathcal{G}$ as a task group that consists of $|\mathcal{G}|$ tasks $\left ( 1\le |\mathcal{G} | \le N \right )$. Then we construct \textit{group-shared MmAP} for CLIP that contains $\mathcal{K}$ transformer layers, including source prompts $P_{\mathcal{G}}=\{P_{\mathcal{G}}^{k}\}_{k=1}^{\mathcal{K}}$, scaling matrices $\mathcal{M}_{\mathcal{G}_{l}}=\{\mathcal{M}_{\mathcal{G}_{l}}^{k}\}_{k=1}^{\mathcal{K}}$ and $\mathcal{M}_{\mathcal{G}_{v}}=\{\mathcal{M}_{\mathcal{G}_{v}}^{k}\}_{k=1}^{\mathcal{K}}$ for language and vision branches, respectively. The \textit{group-shared MmAP} is cumulatively updated by all tasks within group $\mathcal{G}$, achieving complementary benefits across similar tasks. Additionally, for every task in group $\mathcal{G}$, we build \textit{task-specific MmAP} for learning unique task characteristic, including source prompts $P_{\mathcal{T}}=\{P_{\mathcal{T}}^{k}\}_{k=1}^{\mathcal{K}}$, scaling matrices $\mathcal{M}_{\mathcal{T}_{l}}=\{\mathcal{M}_{\mathcal{T}_{l}}^{k}\}_{k=1}^{\mathcal{K}}$ and $\mathcal{M}_{\mathcal{T}_{v}}=\{\mathcal{M}_{\mathcal{T}_{v}}^{k}\}_{k=1}^{\mathcal{K}}$ for language and vision branches. 

During the training of one task $\mathcal{T}$ in group $\mathcal{G}$, we first generate text and image prompts of the $k$-th layers in two encoders, and then we reconstruct the input tokens by composing the class token, the generated prompts and the text/image tokens from the previous layer. Thereby the calculations of the $k$-th layers within text and image encoders can be formally represented as:
\begin{equation}
\begin{aligned}
\left[\_, \_, W_{k}\right]&=\mathcal{L}_{k}\left(\left[P_{\mathcal{G}_l}^k, P_{\mathcal{T}_l}^k,W_{k-1}\right]\right)\\
&=\mathcal{L}_{k}\left(\left[P_{\mathcal{G}}^k \otimes \mathcal{M}_{\mathcal{G}_{l}}^{k},  P_{\mathcal{T}}^k \otimes \mathcal{M}_{\mathcal{T}_{l}}^{k}, W_{k-1}\right]\right),
\end{aligned}
\end{equation}
\begin{equation}
\begin{aligned}
\left[c_{k}, \_, \_, E_{k}\right]&=\mathcal{V}_{k}\left(\left[c_{k-1},P_{\mathcal{G}_v}^k, P_{\mathcal{T}_v}^k,E_{k-1}\right]\right)\\
&=\mathcal{V}_{k}\left(\left[c_{k-1}, P_{\mathcal{G}}^k \otimes \mathcal{M}_{\mathcal{G}_{v}}^{k}, P_{\mathcal{T}}^k \otimes \mathcal{M}_{\mathcal{T}_{v}}^{k}, E_{k-1}\right]\right).
\end{aligned}
\end{equation}

Here $\left [ \cdot , \cdot  \right ] $ refers to the concatenation operation. Finally, \textit{group-shared MmAP} are cumulatively updated by optimizing the following loss:
\begin{equation}
L\left(P_{\mathcal{G}}, \mathcal{M}_{\mathcal{G}_l}, \mathcal{M}_{\mathcal{G}_v}\right)=\sum_{\mathcal{T} \in \mathcal{G}} L_{\mathcal{T}}\left(P_{\mathcal{G}}, \mathcal{M}_{\mathcal{G}_l}, \mathcal{M}_{\mathcal{G}_v}\right),
\end{equation}
and \textit{task-specific MmAP} are trained via:
\begin{equation}
L\left(P_{\mathcal{T}}, \mathcal{M}_{\mathcal{T}_l}, \mathcal{M}_{\mathcal{T}_v}\right)=L_{\mathcal{T}}\left(P_{\mathcal{T}}, \mathcal{M}_{\mathcal{T}_l}, \mathcal{M}_{\mathcal{T}_v}\right),
\end{equation}
where $L_{\mathcal{T}}$ is the cross-entropy loss of task $\mathcal{T}$.

\section{Experiment}
\label{sec:exp}

\subsection{Benchmark Setting}
\paragraph{Datasets.}  
Following prior MTL works~\cite{shen2021variational, long2017learning}, we consider Office-Home~\cite{venkateswara2017deep} and MiniDomainNet~\cite{zhou2021domain} datasets to construct our benchmark. 
\begin{itemize}
    \item \textbf{Office-Home} contains images from four tasks: Art, Clipart, Product and Real World.  Each task covers images from 65 object categories collected under office and home settings. There are about 15,500 images in total. 
    \item \textbf{MiniDomainNet} takes a subset of DomainNet, which is an extremely challenging dataset for multi-task learning. MiniDomainNet has 140,000 images distributed among 126 categories. It contains four different tasks: Clipart, Painting, Sketch and Real.
\end{itemize}

Based on previous research w.r.t MTL and prompt learning \cite{shen2021variational, zhou2022learning}, we randomly select 10\% (6-shot per class) and 20\% (12-shot per class) samples from Office-Home for training, and 1\% (3-shot per class) and 2\% (6-shot per class) samples from MiniDomainNet for training. The remaining samples are reserved for testing.

\paragraph{Baselines.} 
In order to conduct a comprehensive evaluation of our proposed method, we compare it against several tuning baselines, including:
\begin{itemize}
    \item \textbf{Zero-shot} uses hand-crafted text prompt (``a photo of [class]'') templates to zero-shot prediction. 
    \item \textbf{Single-Task Full Fine-Tuning} updates an individual pretrained model for each task and \textbf{Multi-task Full Fine-Tuning} updates an shared pretrained model for all tasks.
    
    \item Single-modal prompt tuning methods, including the standard \textbf{CoOp}~\cite{zhou2022learning}, trained on an individual task for the text prompt tuning, and the standard \textbf{VPT}~\cite{jia2022vpt}, trained on an individual task for the visual prompt tuning. \textbf{CoOp-MT} and \textbf{VPT-MT} are the multi-task version, which train a task-shared prompt with samples from all tasks. Additionally, the recent work \textbf{MaPLe}~\cite{khattak2023maple} serves as one of our baselines, which employs text prompts to generate visual prompts. Similarly, we also construct a multi-task version, referred to as \textbf{MaPLe-MT}. 
    
    \item Other parameter-efficient tuning methods, including \textbf{CLIP-Adapter}~\cite{gao2021clip}, which learns new features on either a visual or a language branch, and \textbf{BitFit}~\cite{zaken2021bitfit}, which tunes the bias parameters of the pre-trained model.
\end{itemize}

\paragraph{Implementation Details.} 
All experiments are conducted using the PyTorch toolkit on NVIDIA V100 GPU, with CLIP (ViT-B/16) chosen as our default model. To ensure a fair comparison, we maintain consistent hyperparameter settings across all parameter efficient tuning methods. Specifically, we use a batch size of 16/4 and train for 5 epochs for Office-Home/MiniDomainNet. We employ the SGD optimizer with a learning rate of 0.0035. We evaluate the checkpoint of the last epoch and run the experiments with three different seeds to obtain the average results. For the source prompt $P_s$, we initialize it with a Gaussian distribution of 0.02 standard deviation. 
\begin{table*}[t]
\centering
\resizebox{1\textwidth}{!}{
    \begin{tabular}{*{14}{cccccccccccccc}}
    \toprule[1pt]
    \midrule[0.3pt]
    
    
    &\multicolumn{1}{c}{}
    & \multicolumn{5}{c}{\textit{Single Task Learning}}   
    & \multicolumn{6}{c}{\textit{Multi Task Learning}}
    \\ 
    \cmidrule(lr){1-1}
    \cmidrule(lr){2-2}
    \cmidrule(lr){3-7}
    \cmidrule(lr){8-14}
    
    & Method
       
    & ZeroShot
    & Full FT
    & CoOp
    & VPT
    & MaPLe

    & Full FT
    & C-Adapter
    & BitFit
    & CoOp-MT
    & VPT-MT
    & MaPLe-MT
    & \textbf{Ours}
    \\ 
    \cmidrule(lr){1-1}
    \cmidrule(lr){2-2}
    \cmidrule(lr){3-7}
    \cmidrule(lr){8-14}
    
    \multirow{5}{*}{10\%} 
    &Art      
    &82.9   
    &84.9$\pm$0.9  &84.2$\pm$0.3   &83.7$\pm$0.5 &84.4$\pm$0.5    
    &\textbf{85.8$\pm$0.8}   &82.3$\pm$0.2   &79.1$\pm$0.1 &84.3$\pm$0.8 &84.0$\pm$0.3 
    &84.8$\pm$0.6
    &\underline{85.7$\pm$0.5}
    \\
    
    &Clipart  
    &68.3  
    &75.4$\pm$0.3  &72.6$\pm$1.1   &70.5$\pm$0.3   &72.8$\pm$0.6
    &\textbf{76.3$\pm$1.0}   &71.7$\pm$0.2   &67.8$\pm$1.4  &73.0$\pm$0.1 &72.4$\pm$0.6
    &73.3$\pm$0.3
    &\textbf{76.3$\pm$0.6}
    \\
    
    &Product  
    &89.3  
    &91.6$\pm$0.3  &92.4$\pm$0.2   &90.9$\pm$0.2   &92.2$\pm$0.1 
    &92.1$\pm$1.3   &90.8$\pm$0.2   &86.7$\pm$1.5  &\underline{92.7$\pm$0.2}  
    &91.7$\pm$0.6
    &\underline{92.7$\pm$0.4}
    &\textbf{92.9$\pm$0.5}
    \\
    
    &Real    
    &90.1  
    &89.8$\pm$0.8  &90.5$\pm$0.3   &89.2$\pm$0.6   &90.4$\pm$0.4
    &90.2$\pm$1.3   &89.2$\pm$0.2   &85.9$\pm$0.5  &90.7$\pm$0.6  &90.6$\pm$0.1
    &\underline{90.8$\pm$0.3}
    &\textbf{90.9$\pm$0.9}
    \\
    
    &Average  
    &82.6  
    &85.4$\pm$0.6  &84.9$\pm$0.5   &83.6$\pm$0.4  &85.0$\pm$0.4  
    &\underline{86.1$\pm$1.1}   &83.5$\pm$0.2   &79.9$\pm$0.9   &85.2$\pm$0.4 &84.7$\pm$0.4
    &85.4$\pm$0.4
    &\textbf{86.5$\pm$0.6}
    \\
    
    \cmidrule(lr){1-1}
    \cmidrule(lr){2-2}
    \cmidrule(lr){3-7}
    \cmidrule(lr){8-14}
    
    \multirow{5}{*}{20\%} 
    &Art
    &84.6     
    &87.1$\pm$1.2  &85.6$\pm$0.6  &85.4$\pm$0.6  &85.9$\pm$0.4
    &\underline{87.4$\pm$0.8}  &83.2$\pm$1.1  &81.7$\pm$0.6  &86.0$\pm$0.2  &85.9$\pm$0.3
    &86.3$\pm$0.4
    &\textbf{88.2$\pm$0.7}
    \\
    
    &Clipart 
    &68.2     
    &77.9$\pm$0.1  &74.5$\pm$0.6  &71.4$\pm$0.4  &74.2$\pm$0.5
    &\textbf{78.8$\pm$1.0}  &75.4$\pm$0.1  &69.6$\pm$1.7  &73.9$\pm$0.6  &72.3$\pm$0.6
    &74.2$\pm$0.6
    &77.1$\pm$0.6\\
    
    &Product 
    &89.5     
    &91.9$\pm$1.3  &\underline{93.0$\pm$0.4}  &91.5$\pm$0.6  &92.8$\pm$0.6
    &\underline{93.0$\pm$1.0}  &91.7$\pm$0.9  &87.2$\pm$1.4  &92.9$\pm$0.4  &92.1$\pm$0.2 
    &92.9$\pm$0.2
    &\textbf{93.5$\pm$0.5}
    \\
    
    &Real    
    &90.7     
    &89.8$\pm$0.6  &91.8$\pm$0.3  &90.9$\pm$0.1 &91.8$\pm$0.4 
    &91.9$\pm$0.4  &90.6$\pm$0.2  &86.7$\pm$1.0  &\underline{92.0$\pm$0.3}  &91.7$\pm$0.3
    &\underline{92.0$\pm$0.5}
    &\textbf{92.4$\pm$0.3}
    \\

    &Average        
    &83.3     
    &86.7$\pm$0.8  &86.2$\pm$0.5  &84.8$\pm$0.4  &86.2$\pm$0.5  
    &\textbf{87.8$\pm$0.8}   &85.2$\pm$0.5   &85.5$\pm$0.4  &86.3$\pm$0.4  &85.5$\pm$0.4
    &86.4$\pm$0.4
    &\textbf{87.8$\pm$0.5}\\
    
    \cmidrule(lr){1-1}
    \cmidrule(lr){2-2}
    \cmidrule(lr){3-7}
    \cmidrule(lr){8-14}
    
    &Parameters      
    &-     
    &598.48 M  
    &0.04 M
    &0.68 M
    &19.2 M
    &149.62 M
    &0.53 M
    &0.17 M
    &0.01 M
    &0.17 M
    &4.8 M
    &0.13 M
    \\

    \midrule[0.3pt]
    \bottomrule
    \end{tabular}
}
\caption{Comparison to various methods on \textit{Office-Home}, using the average accuracy (\%) over 3 different seeds. The benchmark is implemented by us, based on CLIP with ViT-B/16 backbone. We highlight the \textbf{best} and the \underline{second} results.}
\label{tab:office-home_results}
\end{table*}
\begin{table*}[th]
\centering
\resizebox{1\textwidth}{!}{
\setlength\tabcolsep{4pt} 
\begin{tabular}{*{14}{cccccccccccccc}}
\toprule[1pt]\midrule[0.3pt]


&\multicolumn{1}{c}{}
& \multicolumn{5}{c}{\textit{Single Task Learning}}   
& \multicolumn{5}{c}{\textit{Multi Task Learning}}
\\ 
\cmidrule(lr){1-1}
\cmidrule(lr){2-2}
\cmidrule(lr){3-7}
\cmidrule(lr){8-14}

& Method  
& ZeroShot
& Full FT
& CoOp
& VPT
& MaPLe

& Full FT
& C-Adapter
& BitFit
& CoOp-MT
& VPT-MT
& MaPLe-MT
& \textbf{Ours}
\\ 
\cmidrule(lr){1-1}
\cmidrule(lr){2-2}
\cmidrule(lr){3-7}
\cmidrule(lr){8-14}
 
\multirow{5}{*}{1\%} 
&Clipart    
&82.6
&82.1$\pm$1.5 &82.7$\pm$0.1 &82.3$\pm$0.1  &82.9$\pm$0.2  
&82.8$\pm$0.9 &82.6$\pm$0.1 &78.9$\pm$0.4  &\underline{83.4$\pm$0.4} &83.0$\pm$0.3 
&\underline{83.4$\pm$0.4}
&\textbf{83.9$\pm$0.3}
\\

&Painting   
&82.3   
&81.8$\pm$0.7 &81.8$\pm$0.3   &81.7$\pm$0.3  &82.0$\pm$0.2  
&81.5$\pm$0.6 &80.4$\pm$0.3   &74.7$\pm$0.4 &82.3$\pm$0.2 &81.9$\pm$0.6 
&\underline{82.5$\pm$0.4}
&\textbf{83.5$\pm$0.2}
\\

&Real       
&91.2   
&89.1$\pm$0.5 &91.9$\pm$0.3   &91.6$\pm$0.2   &\underline{92.0$\pm$0.3}
&89.1$\pm$0.6   &90.9$\pm$0.2   &84.2$\pm$0.3 &91.3$\pm$0.1  &90.1$\pm$0.2
&91.4$\pm$0.1
&\textbf{92.2$\pm$0.2}
\\

&Sketch     
&79.9   
&77.0$\pm$0.7  &77.1$\pm$0.2   &78.5$\pm$0.3  &78.5$\pm$0.4
&77.2$\pm$1.0   &78.3$\pm$0.6   &72.4$\pm$0.5 &\underline{79.2$\pm$0.2} &78.6$\pm$0.6
&79.1$\pm$0.2
&\textbf{79.8$\pm$0.7}
\\

&Average    
&84.0   
&82.5$\pm$0.9  &83.4$\pm$0.2   &83.5$\pm$0.2   &83.9$\pm$0.3  
&82.7$\pm$0.8   &83.0$\pm$0.3   &77.6$\pm$0.4  &84.0$\pm$0.3 &83.4$\pm$0.4   &\underline{84.1$\pm$0.3} 
&\textbf{84.9$\pm$0.4}
\\

\cmidrule(lr){1-1}
\cmidrule(lr){2-2}
\cmidrule(lr){3-7}
\cmidrule(lr){8-14}

\multirow{5}{*}{2\%} 
&Clipart    
&82.6   
&82.2$\pm$1.3   &83.8$\pm$0.1  &83.5$\pm$0.3  &83.8$\pm$0.4
&82.8$\pm$0.9   &83.1$\pm$0.1  &81.5$\pm$0.2  &\underline{84.7$\pm$0.3}  &83.8$\pm$0.5 
&84.5$\pm$0.3
&\textbf{85.7$\pm$0.4}
\\
&Painting   
&82.3    
&82.1$\pm$1.4   &82.5$\pm$0.2  &82.4$\pm$0.1  &82.7$\pm$0.2
&82.1$\pm$0.7   &81.5$\pm$0.3  &76.8$\pm$0.2  &83.2$\pm$0.2 &82.2$\pm$0.1   
&\underline{83.6$\pm$0.4}
&\textbf{85.0$\pm$0.2}
\\
&Real       
&91.2    
&89.2$\pm$0.8   &\underline{91.9$\pm$0.1}  &91.5$\pm$0.1   &91.6$\pm$0.2
&89.3$\pm$0.5   &90.6$\pm$0.2  &85.9$\pm$0.1   &91.7$\pm$0.1 &90.5$\pm$0.1  
&\underline{91.9$\pm$0.2}
&\textbf{92.3$\pm$0.1}
\\
&Sketch     
&80.0    
&77.4$\pm$0.5   &79.0$\pm$0.5  &79.6$\pm$0.2  &79.9$\pm$0.3
&77.7$\pm$1.0   &78.7$\pm$0.6  &74.8$\pm$0.2  &80.1$\pm$0.4 &79.0$\pm$0.2
&\underline{80.5$\pm$0.3}
&\textbf{81.5$\pm$0.2}
\\

&Average    
&84.0    
&82.7$\pm$1.0   &84.3$\pm$0.2  &84.2$\pm$0.2 &84.5$\pm$0.3
&83.0$\pm$0.8   &83.4$\pm$0.3  &79.8$\pm$0.2 &84.9$\pm$0.4 &83.9$\pm$0.3
&\underline{85.1$\pm$0.3}
&\textbf{86.1$\pm$0.2}
\\

\cmidrule(lr){1-1}
\cmidrule(lr){2-2}
\cmidrule(lr){3-7}
\cmidrule(lr){8-14}

&Parameters     
&-     
&598.48 M  
&0.04 M
&0.68 M
&19.2 M
&149.62 M
&0.53 M
&0.17 M
&0.01 M
&0.17 M
&4.8 M
&0.13 M
\\

\midrule[0.3pt]
\bottomrule
\end{tabular}
}

\caption{Comparison to various methods on \textit{MiniDomainNet}, using the average accuracy (\%) over 3 different seeds. The benchmark is implemented by us, based on CLIP with ViT-B/16 backbone. We highlight the \textbf{best} and the \underline{second} results.}
\label{tab:domainnet_results}
\end{table*}
\subsection{Experiment Results}
\paragraph{Office-Home.} 
The results are presented in Table~\ref{tab:office-home_results}. Firstly, we observe that our method is on par with Multi-Task Full Fine-Tuning across different data splits (10\% or 20\%) while requiring only 0.09\% (0.13M vs. 149.62M) trainable parameters. This represents a significant breakthrough in parameter-efficient tuning of CLIP for multi-task image recognition. Secondly, our method consistently outperforms other parameter efficient tuning methods. In comparison to prompt methods (i.e., MaPLe-MT, CoOp-MT, and VPT-MT), our method exhibits a significant improvement, highlighting the necessity of integrating visual and text modalities when tuning CLIP and combining the group-shared and the task-specific knowledge. 

Regarding the number of trainable parameters, our method ranks second only to CoOp-MT, achieving the best trade-off between accuracy and trainable parameters. Thirdly, we also find that prompt methods outperform CLIP-Adapter and BitFit, indicating that aligning downstream data with CLIP is a more efficient approach.

\paragraph{MiniDomainNet.}
The results are shown in Table~\ref{tab:domainnet_results}. We can draw consistent conclusions with Office-Home. Our method performs the best and achieves 84.9\% on the 1\% split and 86.1\% on the 2\% split. However, we observe that the performance of Full Fine-Tuning is not very satisfactory and is worse than most parameter-efficient tuning methods, which is caused by overfitting. Specifically, the task difficulty of MiniDomainNet is significantly increased compared to Office-Home, and concurrently, the number of training data is limited. Moreover, the BitFit method exhibits the worst performance. It updates few parameters of the CLIP using a small amount of data, which severely impairs the original zero-shot capability of CLIP. 

The effects of CoOp-MT, VPT-MT, and MaPLe-MT can only approach zero-shot on the 1\% split, but when the amount of training data reaches 2\%, CoOp-MT and MaPLe-MT surpass zero-shot by 0.9\% and 1.1\%, respectively. Therefore, to explore the performance under different training data sizes, we set up related experiments, as detailed in ablation study.

\subsection{Ablation Study} In this section, we construct various ablation experiments to further analyze our proposed MmAP and multi-task prompt learning framework. At the same time, we also design related experiments for different downstream data size.
\begin{figure}[h]
   \begin{picture}(0,164)
     \put(0,-4){\includegraphics[width=1\linewidth]{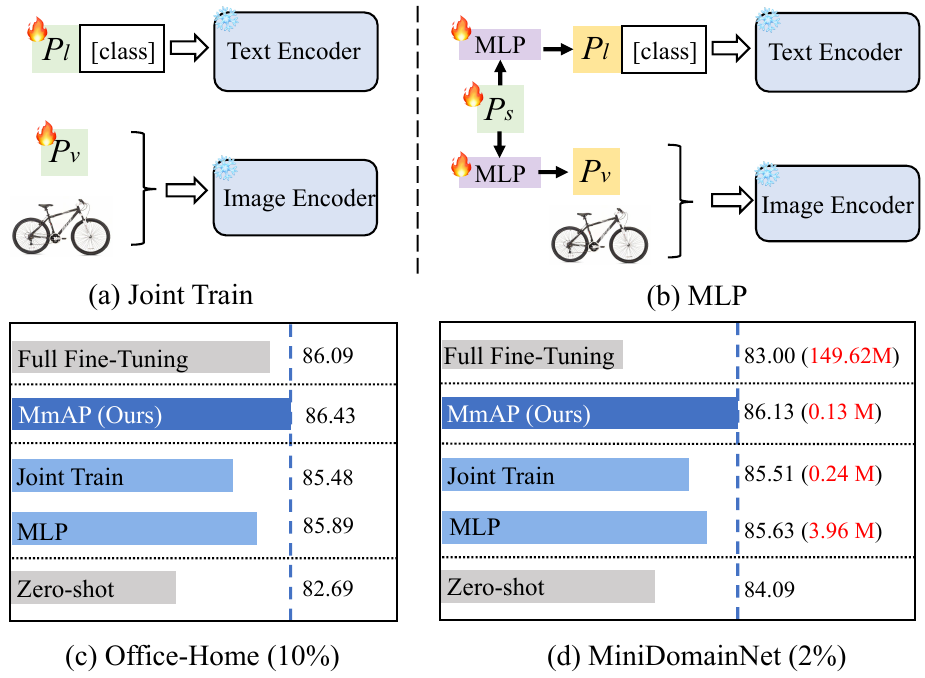}}
   \end{picture}
    \caption{Ablation study of MmAP on Office-Home and miniDomainNet datasets. We construct two baselines: (a) jointly training the text and visual prompts, and (b) utilizing two MLP layers to generate the text and visual prompts.}
    \label{fig:ablation_MmAP}
\end{figure}
\begin{figure*}[t]
	\includegraphics[width=1\textwidth]{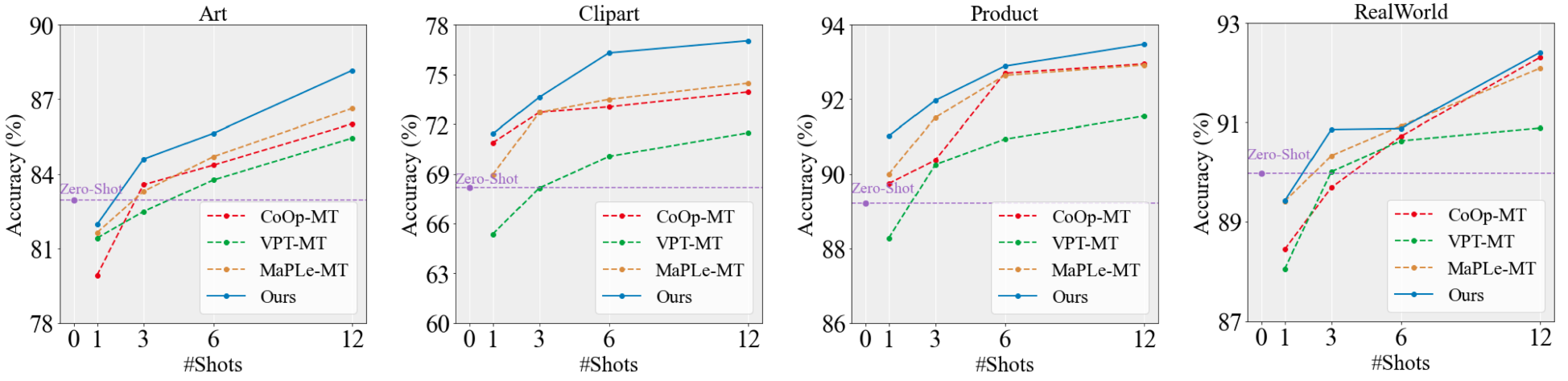}
	\caption{Main results on Office-Home (four tasks) under the k-shot setting. We report the accuracy (\%) for 1/3/6/12 shots. Overall, our method attains substantial improvements over zero-shot CLIP and performs favorably against other baselines.}
	\label{fig:shot}
\end{figure*}
\paragraph{Effectiveness of MmAP.} To verify the effectiveness of our proposed MmAP, we set up related ablation experiments. As displayed in Figure~\ref{fig:ablation_MmAP}a, a straightforward approach for multi-modal prompts is to tune the text and visual prompts jointly. Another straightforward solution involves sharing text and visual prompts. However, since the dimensions of the Text Encoder ($d_l=512$) and Image Encoder ($d_v=768$) of CLIP are not equal, they cannot be shared directly. Therefore, we design the MLP prompt baseline as another comparison scheme, which employs two MLP layers to generate the text and visual prompts, as shown in~\ref{fig:ablation_MmAP}b.

The results are shown in Figure~\ref{fig:ablation_MmAP}c. Across the four tasks of Office-Home, the MLP baseline exhibits a 0.5\% improvement compared to the joint train baseline, demonstrating the effectiveness of establishing a connection between the text prompt and the visual prompt. Additionally, we observe that MmAP achieves a 0.54\% improvement compared to the MLP baseline, indicating that the MmAP method is more effective in maximizing information sharing between the text and visual prompts through the Kronecker Product. At the same time, MmAP trainable parameters are greatly reduced relative to the MLP baseline (0.13M vs. 3.96M).

\begin{table}[h]
    \centering
    \resizebox{0.45\textwidth}{!}{
    \begin{tabular}{ccccc}
    \toprule
    \multirow{2}{*}{Task Specific} & \multicolumn{2}{c}{Group Shared}          & \multicolumn{2}{c}{Office-Home} \\
    \cmidrule(lr){2-3}\cmidrule(lr){4-5}
    & Task Group             & Random            & 10\%           & 20\%           \\ \midrule
    \checkmark                              & \usym{2613}                      & \usym{2613}                 & 85.76          & 86.97          \\
    \usym{2613}                              & \checkmark                      & \usym{2613}                 & 86.05          & 87.29          \\
    \checkmark                              & \usym{2613}                      & \checkmark                 & 85.80          & 86.92          \\
    \checkmark                              & \checkmark                      & \usym{2613}                 & \textbf{86.48} & \textbf{87.77} \\
    \checkmark                              & \multicolumn{2}{c}{\textit{All in one group}} & 86.09          & 87.36          \\ \bottomrule
    \end{tabular}
}
    \caption{Ablation study of Multi-Task Prompt Learning Framework. ``Random'' means grouping tasks randomly.} 
\label{tab:ablation_framework}
\end{table}
\paragraph{Effectiveness of Multi-Task Prompt Learning Framework.}
In our multi-task prompt learning framework, task-specific MmAP and group-shared MmAP are the primary components. To verify the importance of each module, we conduct related ablation experiments on Office-Home, and the results are presented in Table~\ref{tab:ablation_framework}. To substantiate the effectiveness of task grouping strategy, we incorporate random grouping as a benchmark for comparison. The empirical results elucidate that each module within our framework plays a pivotal role, cumulatively contributing to the superior performance achieved by our multi-task prompt learning framework. Compared to the random grouping, our task grouping performs 0.68\% and 0.85\% higher under the settings of 10\% and 20\%, respectively. Compared to the all task in one group, our task grouping performs 0.39\% and 0.41\% higher under the settings of 10\% and 20\%. From another perspective, task-specific MmAP surpasses that of CoOp and VPT (results in Table~\ref{tab:office-home_results}), further demonstrating the effectiveness of our MmAP.

\paragraph{Different Downstream Data Size.}
\label{sec:different_data_size}
We examine the impact of training data size on Office-Home (four tasks). We select 1/3/6/12 shots per class and compare our MmAP with CoOp-MT, VPT-MT, and MaPLe-MT. The results for each task and method at different training data scales are presented in Figure~\ref{fig:shot}. The results indicate that our method surpasses all other baselines on the four tasks across data scales, confirming our method's strong generalization. However, we observe that all methods underperform in comparison to Zero-Shot in the 1-shot setting for Art and Real World tasks. This may be due to the fact that 1-shot is too specific to serve as a general representation for the entire task. When provided with 3 or more shots for training, the average performance gap introduced by our method is substantial.

\section{Conclusion}
\label{sec:conclusion}
In this work, we propose the Multi-modal Alignment Prompt (MmAP) for adapting CLIP to downstream tasks, which achieves the best trade-off between trainable parameters and performance against most of the existing methods. Simultaneously, MmAP addresses the issue of previous single-modal prompt methods (e.g., CoOp and VPT) disrupting CLIP's modal alignment. Building on MmAP, we design a multi-task prompt learning framework, which not only enables similar tasks to be trained together to enhance task complementarity but also preserves the independent characteristics of each task. Our approach achieves significant performance improvements compared to full fine-tuning on two large multi-task learning datasets under limited downstream data while only utilizing $\sim0.09\%$ trainable parameters.

\bibliography{ref}

\end{document}